\documentclass[11pt,a4paper]{article}
\usepackage[hyperref]{emnlp2018}
\usepackage{times}
\usepackage{latexsym}
\usepackage{qtree}
\usepackage{graphicx}
\usepackage{amsmath}
\usepackage{amsfonts,amssymb}
\usepackage{multirow}
\usepackage{multicol}
\usepackage{enumitem}
\usepackage{comment}
\usepackage{url}
\usepackage{afterpage}

\aclfinalcopy 
\def\aclpaperid{1858} 

\newcommand{\figref}[1]{Figure \ref{#1}}
\newcommand{\tabref}[1]{Table \ref{#1}}
\newcommand{\secref}[1]{Section \ref{#1}}
\newcommand{\equaref}[1]{Equation \eqref{#1}}

\title{Toward Fast and Accurate Neural Discourse Segmentation}

\author{Yizhong Wang \qquad
  Sujian Li \qquad Jingfeng Yang\\
 MOE Key Lab of Computational Linguistics, School of EECS, Peking University  \\
  {\tt \{yizhong, lisujian, yjfllpyym\}@pku.edu.cn} \\}

\date{}

\begin{document}
\maketitle
\begin{abstract}
Discourse segmentation, which segments texts into Elementary Discourse Units, is a fundamental step in discourse analysis. Previous discourse segmenters rely on complicated hand-crafted features and are not practical in actual use. In this paper, we propose an end-to-end neural segmenter based on BiLSTM-CRF framework. To improve its accuracy, we address the problem of data insufficiency by transferring a word representation model that is trained on a large corpus. We also propose a restricted self-attention mechanism in order to capture useful information within a neighborhood. Experiments on the RST-DT corpus show that our model is significantly faster than previous methods, while achieving new state-of-the-art performance.
\footnote{Our code is available at \url{https://github.com/PKU-TANGENT/NeuralEDUSeg}}
\end{abstract}

\section{Introduction}
\label{sec:introduction}

Discourse segmentation, which divides text into proper discourse units, is one of the fundamental tasks in natural language processing. According to Rhetorical Structure Theory (RST) \citep{mann1988rhetorical}, a complex text is composed of non-overlapping Elementary Discourse Units (EDUs), as shown in \tabref{tab:example}. Segmenting text into such discourse units is a key step in discourse analysis \citep{marcu2000theory} and can benefit many downstream tasks, such as sentence compression \citep{sporleder2005discourse} or document summarization \citep{junyi2016summarization}.

Since EDUs are initially designed to be determined with lexical and syntactic clues \citep{rst-dt}, existing methods for discourse segmentation usually design hand-crafted features to capture these clues \cite{two-pass}. Especially, nearly all previous methods rely on syntactic parse trees to achieve good performance. But extracting such features usually takes a long time, which contradicts the fundamental role of discourse segmentation and hinders its actual use. 
Considering the great success of deep learning on many NLP tasks \citep{DBLP:conf/naacl/LuL16}, 
it's a natural idea for us to design an end-to-end neural model that can segment texts fast and accurately.



\begin{table}[tb]
\centering
\begin{tabular}{|p{0.45\textwidth}|}
\hline
\textbf{[}Mr. Rambo says\textbf{]}$_{e_1}$ \textbf{[}that a 3.2-acre property\textbf{]}$_{e_2}$ \textbf{[}overlooking the San Fernando Valley\textbf{]}$_{e_3}$ \textbf{[}is priced at \$4 million\textbf{]}$_{e_4}$ \textbf{[}because the late actor Erroll Flynn once lived there.\textbf{]}$_{e_5}$ \\
\hline
\end{tabular}
\caption{A sentence that is segmented into five EDUs} \label{tab:example}
\end{table}

The first challenge of applying neural methods to discourse segmentation is data insufficiency. Due to the limited size of labeled data in existing corpus \citep{rst-dt}, it's quite hard to train a data-hungry neural model without any prior knowledge. 
In fact, some traditional features, such as the POS tags or parse trees, naturally provide strong signals for identifying EDUs. Removing them definitely increases the difficulty of learning an accurate model.  
Secondly, many EDU boundaries are actually not determined locally. For example, to recognize the boundary between $e_3$ and $e_4$ in \tabref{tab:example}, our model has to be aware that $e_3$ is an embedded clauses starting from ``overlooking'', otherwise it could regard ``San Fernando Valley'' as the subject of $e_4$. Such kind of long-distance dependency can be precisely  extracted from parse trees but is difficult for neural models to capture. 

To address these challenges, in this paper, we propose a neural discourse segmenter based on the BiLSTM-CRF \cite{lstm-crf} framework and further improve it from two aspects. Firstly, since the discourse segmentation corpus is too small to learn precise word representations, we transfer a word representation model trained on a large corpus into our task, and show that this transferred model can provide very useful information for our task. Secondly, in order to model long-distance dependency, we employ the self-attention mechanism \cite{transformer} when encoding the text. Different from previous self-attention, we restrict the attention area to a neighborhood of fixed size.
The motivation is that effective information for determining the boundaries is usually collected from adjacent EDUs, while the whole text may contain many disturbing words, which could mislead the model into incorrect decisions. 
In summary, the contributions of this work are as follows:
\begin{itemize}
\item Our neural discourse segmentation model doesn't rely on any syntactic features, while it can outperform other state-of-the-art systems and achieve significant speedup.
\item To our knowledge, we are the first to transfer word representations learned from large corpus into discourse segmentation task and show that they can significantly alleviate the data insufficiency problem.
\item Based on the nature of discourse segmentation, we propose a restricted attention mechanism , which enables the model to capture useful information within a neighborhood but ignore unnecessary faraway noises. 
\end{itemize}
\section{Neural Discourse Segmentation Model}
\label{sec:method}
\begin{figure}[tb]
\centering
\includegraphics[width=0.48\textwidth]{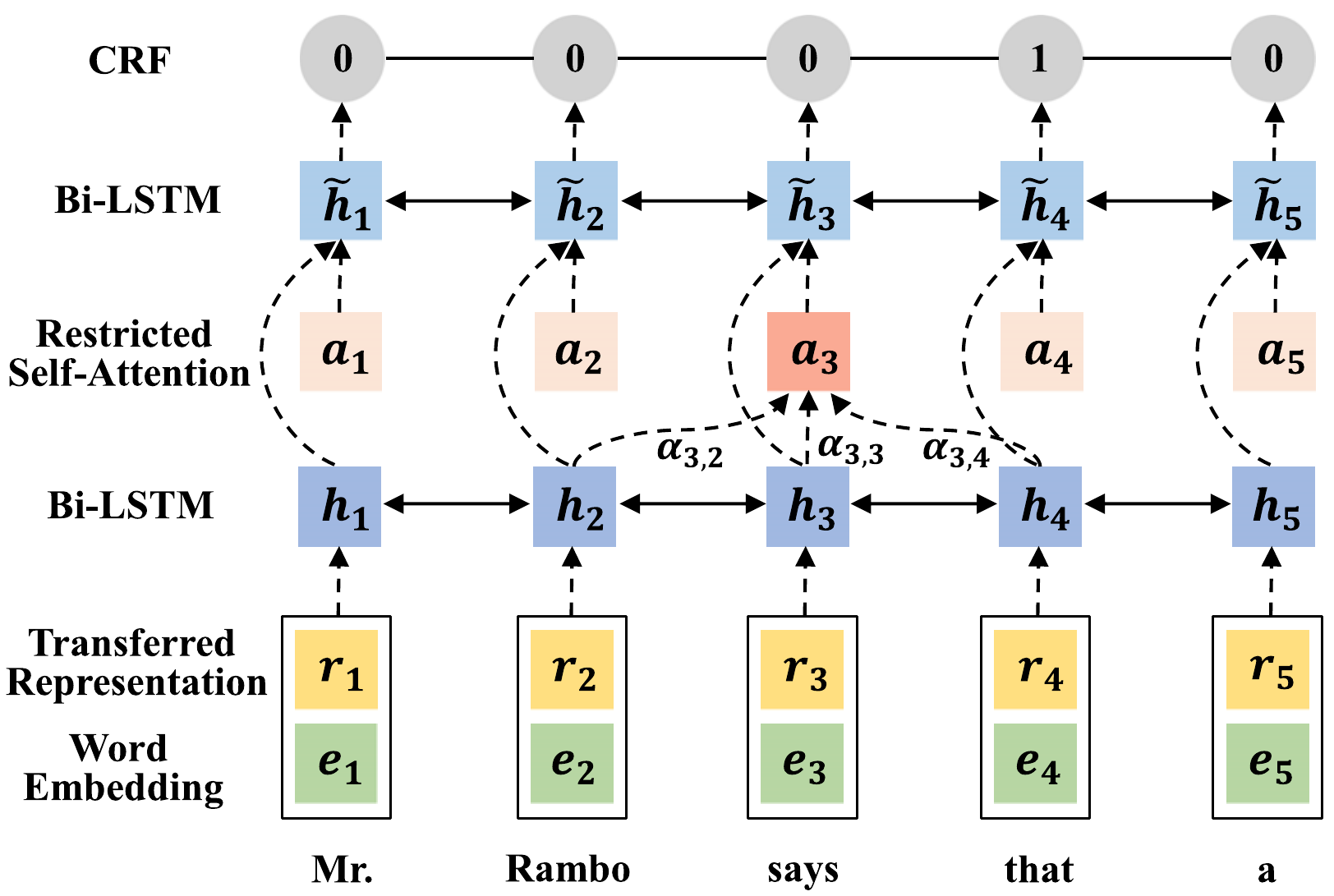}
\caption{Overview of our model for discourse segmentation}
\label{fig:architecture}
\end{figure}

We model discourse segmentation as a sequence labeling task, where the start word of each EDU (except the first EDU) is supposed to be labeled as 1 and other words are labeled as 0. \figref{fig:architecture} gives an overview of our segmentation model. We will introduce the BiLSTM-CRF framework in \secref{sec:lstm-crf}, and describe the two key components of our model in \secref{sec:transfer}, \ref{sec:attention}.

\subsection{BiLSTM-CRF for Sequence Labeling}
\label{sec:lstm-crf}

Conditional Random Fields (CRF) \cite{crf} is an effective method to sequence labeling problem and has been widely used in many NLP tasks \cite{DBLP:journals/ftml/SuttonM12}. To approach our discourse segmentation task in a neural way, we adopt the BiLSTM-CRF model \cite{lstm-crf} as the framework of our system. Formally, given an input sentence $\mathbf{x}=\{x_t\}_{t=1}^{n}$, we first embed each word into a vector $\mathbf{e}_t$. Then these word embeddings are fed into a bi-directional LSTM layer to model the sequential information:
\begin{align}
\mathbf{h}_t &= \textrm{BiLSTM}(\mathbf{h}_{t-1}, \mathbf{e}_t)
\label{eq:bi-lstm}
\end{align}
where $\mathbf{h}_t$ is the concatenation of the hidden states from both forward and backward LSTMs. After encoding this sentence, we make labeling decisions for each word. Instead of modeling the decisions independently, the CRF layer computes the conditional probability $p(\mathbf{y}|\mathbf{h}; \mathbf{W}, \mathbf{b})$ over all possible label sequences $\mathbf{y}$ given $\mathbf{h}$ as follows:
\begin{align}
p(\mathbf{y}|\mathbf{h}; \mathbf{W}, \mathbf{b}) = \frac{\prod_{i=1}^{n} \psi_i(y_{i-1}, y_i, \mathbf{h})}{\sum\limits_{{y'}\in\mathcal{Y}}\prod_{i=1}^{n} \psi_i(y'_{i-1}, y'_i, \mathbf{h})}
\end{align}
where $\psi_i(y_{i-1}, y_i, \mathbf{h}) = \textrm{exp}(\mathbf{w}^T\mathbf{h}_i+b)$ is the potential function and $\mathcal{Y}$ is the set of possible label sequences. The training objective is to maximize the conditional likelihood of the golden label sequence. During testing, we search for the label sequence with the highest conditional probability.

\subsection{Transferring Representations Learned from Large Corpus}
\label{sec:transfer}

Due to the large parameter space, neural models usually require much more training data in order to achieve good performance. However, to the best of our knowledge, nearly all existing discourse segmentation corpora are limited in size. After we remove all the syntactic features, which has been proven useful in many previous work \cite{bach2012reranking,two-pass, joty15codra}, it's expected that our neural model will not achieve very satisfying results. 

To tackle this issue, we propose to leverage model learned from other large datasets, aiming that this transferred model has been well trained to encode text and capture useful signals. Instead of training the transferred model by ourselves, in this paper, we adopt the ELMo word representations \cite{elmo}, which are derived from a bidirectional language model (BiLM) trained on one billion word benchmark corpus \cite{one-billion}. Specifically, this BiLM has one character convolution layer and two biLSTM layers, and correspondingly there are  three internal representations for each word $x_t$, which are denoted as $\{\mathbf{h}_{t,l}^\textrm{LM}\}_{l=1}^{3}$. Following \cite{elmo}, we compute the ELMo representation $\mathbf{r}_t$ for word $x_t$ as follows:
\begin{align}
\mathbf{r}_t = \gamma^{\textrm{LM}} \sum\nolimits_{l=0}^{3} s_l^{\textrm{LM}} \mathbf{h}_{t,l}^\textrm{LM}
\end{align}
where $\mathbf{s}^{\textrm{LM}}$ are normalized weights and $\gamma^{\textrm{LM}}$ controls the scaling of the entire ELMo vector. Then we concatenate $\mathbf{r}_t$ with the word embedding $\mathbf{e}_t$, and take them as the input of \equaref{eq:bi-lstm}.

\subsection{Restricted Self-Attention}
\label{sec:attention}

As we have introduced in \secref{sec:introduction}, some EDU boundaries rely on relatively long-distance signals to recognize, while normal LSTM model is still weak at this. Recently, self-attention mechanism, which relates different positions of a single sequence, has been successfully applied to many NLP tasks \cite{transformer,rnet} and shows its superiority in capturing long dependency. However, we found that most boundaries are actually only influenced by nearby EDUs,  thereby forcing the model to attend to the whole sequence will bring in unnecessary noises. Therefore, we propose a restricted self-attention mechanism, which only collects information from a fixed neighborhood. To do this, we first compute the similarity between current word $x_i$ and each nearby word $x_j$ within a window:
\begin{align}
s_{i, j} = \mathbf{w}_{attn}^{T}[\mathbf{h}_i, \mathbf{h}_j, \mathbf{h}_i \odot \mathbf{h}_j]
\end{align}

Then the attention vector $\mathbf{a}_i$ is computed as a weighted sum of nearby words:
\begin{align}
\alpha_{i, j} &=  {\frac {e^{s_{i, j}}}{\sum_{k=-K}^{K}e^{s_{i,i+k}}}} \\
\mathbf{a}_{i} &= \sum\nolimits_{j=-K}^{K} \alpha_{i, i+k}\mathbf{h}_{i+k} 
\end{align}
where hyper-parameter $K$ is the window size. This attention vector $\mathbf{a}_{i}$ is then put into another BiLSTM layer together with $\mathbf{h}_i$ in order to fuse the information:
\begin{align}
\mathbf{\tilde{h}}_t &= \textrm{BiLSTM}(\mathbf{\tilde{h}}_{t-1}, [\mathbf{h}_t, \mathbf{a}_t])
\end{align}

We use $\mathbf{\tilde{h}}_t$ as the new input to the CRF layer.
\section{Experiments and Results}

\subsection{Dataset and Metrics}
We conduct experiments on the RST Discourse Treebank (RST-DT) \citep{rst-dt}. The original corpus contains 385 Wall Street Journal articles from the Penn Treebank, which are divided in to training set (347 articles, 6132 sentences) and test set (38 articles, 991 sentences). We randomly sample 34 (10\%) articles from the train set as validation set in order to tune the hyper-parameters and only train our model on the remained train set. We follow mainstream studies \citep{soricut2003sentence,joty15codra} to measure segmentation accuracy only with respect to the intra-sentential segment boundaries, and we report Precision (P), Recall (R) and F1-score (F1) for segmentation performance.

\subsection{Implementation Details}
We tune all the hyper-parameters according to the
model performance on the separated validation set. The 300-D Glove embeddings \citep{pennington2014glove} are employed and kept fixed during training. We use the AllenNLP toolkit \citep{gardner2018allennlp} to compute the ELMo word representations. The hidden size of our model is set to be 200 and the batch size is 32. L2 regularization is applied to trainable variables with its weight as
0.0001 and we use dropout between every two layers, where the dropout rate is 0.1. For model training, we employ the Adam algorithm \citep{adam} with its initial
learning rate as 0.0001 and we clip the gradients to a maximal norm 5.0. Exponential moving average is applied to all trainable variables with a decay rate 0.9999. The window size $K$ for restricted attention is set to be 5. 

\subsection{Performance}

\afterpage{
  \begin{table}[tbp]
  \centering
  \begin{tabular}{c|c|c|c|c}
  \hline
  Model        & Tree & P(\%) & R(\%) & F1(\%) \\ 
  \hline
  SPADE   & Gold & 84.1   & 85.4  & 84.7 \\
  NNDS    & Gold & 85.5   & 86.6  & 86.0 \\
  CRFSeg  & Gold & 92.7  & 89.7  & 91.2 \\
  Reranking & Gold &  \textbf{93.1} & 94.2  & 93.7 \\
  \hline \hline
  CRFSeg   & Stanford  & 91.0   & 87.2  & 89.0 \\ 
  CODRA & BLLIP & 88.0 & 92.3 & 90.1 \\
  Reranking   & Stanford  & 91.5   & 90.4 &  91.0\\ 
  Two-Pass    & BLLIP      &   92.8 & 92.3 & 92.6 \\
  \hline \hline
  Our Model & No  & 92.9  & \textbf{95.7} & \textbf{94.3} \\ 
  - Attention & No  &  92.4  & 94.8 & 93.6 \\ 
  - ELMo & No  &  87.9  & 84.5 & 86.2 \\ 
  - Both & No  & 87.0  & 82.8 & 84.8 \\ 
  \hline
  Human          & No & 98.5 & 98.2 & 98.3\\ \hline
  \end{tabular}
  \caption{Performance of our model and other systems on the RST-DT test set \footnotemark}
  \label{tab:performance}
  \end{table}
\footnotetext{In parallel with our work, \newcite{DBLP:conf/ijcai/LiSJ18} proposes another neural model with its performance as: P-91.6,
R-92.8,
F1-92.2. We didn't see their paper at the time of submission, but it's worth mentioning here for the readers' reference.}
}
The results of our model and other competing systems on the test set of RST-DT are shown in \tabref{tab:performance}. We compare our results against the following systems: (1) \textbf{SPADE} \citep{soricut2003sentence} is an early system using simple lexical and syntactic features; (2) \textbf{NNDS} \citep{subba2007automatic} uses a neural network classifier to do the segmentation after extracting features; (3) \textbf{CRFSeg} \citep{hernault2010hilda} is the first discourse segmenter using CRF model; (4) \textbf{CODRA} \citep{joty15codra} uses fewer features and a simple logistic regression model to achieve impressive results; (5) \textbf{Reranking} \citep{bach2012reranking} reranks the N-best outputs of a base CRF segmenter; (6) \textbf{Two-Pass} \citep{two-pass} conducts a second segmentation after extracting global features from the first segmentation result. 
All these methods rely on tree features and we list their performance given different parse trees, where \textbf{Gold} are the trees extracted from the Penn Treebank \citep{prasad2005penn}, \textbf{Stanford} represents trees from the Stanford parser \cite{stanford-parsing} and \textbf{BLLIP} represents those from the BLLIP parser \citep{bllip}. It should be noted that the results of SPADE and CRFSeg are taken from \newcite{bach2012reranking} since the original papers adopt different evaluation metrics. All the other results are taken from the corresponding original papers.

From \tabref{tab:performance}, we can see that our model achieves state-of-the-art performance without extra parse trees. Especially, if no gold parse trees are provided, our system outperforms other methods by more than 1.7 points in F1 score. Since the gold parse trees are not available when processing new sentences, this improvement becomes more valuable when the system is put into use.

To further explore the influence of different components in our model, we also report the results of ablation experiments in \tabref{tab:performance}. We can see that the transferred ELMo representations provide the most significant improvement. This accords with our assumption that the RST-DT corpus itself is not large enough to train an expressive neural model sufficiently. With the help of the transferred representations, we are capable of capturing more semantic and syntactic signals. Also, comparing the models with and without the restricted self-attention, we find that this attention mechanism can further boost the performance. Especially, if there are no ELMo vectors, the improvement provided by the attention mechanism is more noticeable.

\subsection{Speed Comparison}

\begin{table}[tbp]
\centering
\begin{tabular}{c|c|c}
\hline
System & Speed (Sents/s)& Speedup \\
\hline
Two-Pass & 1.39 & 1.0x\\
SPADE  & 3.78 & 2.7x \\
\hline
Ours (Batch=1) & 9.09  &  6.5x \\
Ours (Batch=32) & 76.23  &  54.8x \\
\hline
\end{tabular}
\caption{Speed comparison with two open-sourced discourse segmenter}
\label{tab:speed} 
\end{table}

We also measure the speedup of our model against traditional systems in \tabref{tab:speed}. The \textbf{Two-Pass} system has the best performance among all existing methods, while \textbf{SPADE} is much simpler with less features. We test these systems on the same machine (CPU: Intel Xeon E5-2690, GPU: NVIDIA Tesla P100). The results show that our system is 2.4-6.5 times faster than the compared systems if the batch size is 1. Moreover, if we process the test sentences in parallel, we can achieve 20.2-54.8 times speedup with the batch size as 32. This makes our system more practical in actually use.

\subsection{Effect of Restricted Self-Attention}

We propose to restrict the self-attention within a neighborhood instead of the whole sequence. 
\tabref{tab:window} demonstrates the performance of our model over different window size $K$. We can see that all these results is better than the performance our model without attention mechanism. However, a proper restriction window is helpful for the attention mechanism to take better effect. 

\begin{table}[htbp]

\centering
\begin{tabular}{|c|c|c|c|c|c|c|}
\hline
Window Size & 1 & 5 & 10 & $\infty$ \\
\hline
F1-score & 94.0 & 94.3 & 94.2 & 93.8 \\
\hline
\end{tabular}
\caption{Performance of our model over different attention window size}
\label{tab:window} 
\end{table}

\section{Conclusion}

In this paper, we propose a neural discourse segmenter that can segment text fast and accurately. Different from previous methods, our segmenter doesn't rely on any hand-crafted features, especially the syntactic parse tree. To achieve our goal, we propose to leverage the word representations learned from large corpus and we also propose a restricted self-attention mechanism. Experimental results on RST-DT show that our system can achieve state-of-the-art performance together with significant speedup.

\section*{Acknowledgments}
We thank the anonymous reviewers for their insightful comments on this paper.
This work was partially supported by National Natural Science Foundation of China (61572049 and 6187022165).
The corresponding author of this paper is Sujian Li.

\bibliography{emnlp2018}
\bibliographystyle{acl_natbib_nourl}

\end{document}